\pdfoutput=1
 
\documentclass[journal]{IEEEtran}
 
\usepackage{amsmath}
\usepackage{amssymb}
\usepackage{graphicx}
\usepackage{cite}
\usepackage{array}
\usepackage{multirow}
\usepackage{booktabs}
\usepackage{algorithmic}
\usepackage{textcomp}
\usepackage{xcolor}
\usepackage{url}
\graphicspath{{figures/}}
 
\begin{document}
 
\title{GM-GReFEL: Geometry-Guided Spatiotemporal State-Space Learning for Robust Dynamic Facial Expression Recognition}
 
\author{Yudhistira~Arditya~Pratama and Yi-Zeng~Hsieh,~\IEEEmembership{Senior Member,~IEEE}
\thanks{Y.~A.~Pratama and Y.-Z.~Hsieh are with the Department of Electrical Engineering, National Taiwan University of Science and Technology, Taipei 106335, Taiwan (e-mail: M11307808@mail.ntust.edu.tw; yzhsieh@mail.ntust.edu.tw). Corresponding author: Yi-Zeng Hsieh.}
\thanks{This work has been submitted to the IEEE for possible publication. Copyright may be transferred without notice, after which this version may no longer be accessible.}}
 
\markboth{}%
{Pratama \MakeLowercase{\textit{et al.}}: GM-GReFEL: Geometry-Aware Spatiotemporal State-Space Architecture for DFER in the Wild}
 
\IEEEtitleabstractindextext{
\begin{abstract}
Recognizing facial expressions in unconstrained video suffers from two coupled problems: self-attention grows quadratically in cost with clip length, and crowdsourced labels are noisy. We present GM-GReFEL, a spatiotemporal state-space network with a geometry-aware decision stage. Its Modulated GroupMamba backbone replaces a spatiotemporal Vision Transformer. Six channel groups are scanned separately, four over the frame plane and two over time, so the cost grows linearly with sequence length ($\mathcal{O}(N)$); after each layer, a Channel Affinity Modulation gate mixes the groups again. On top of the backbone, a GReFEL reliability module turns the normalized entropy of each prediction into a mixing weight between the classifier output and a vote from trainable geometric anchors, so that facial geometry counts for more when the network is unsure. The model is initialized by static-to-dynamic weight transfer and trained with a Triple Loss. In five-fold cross-validation, the best fold scores 66.57\%/76.46\% UAR/WAR on DFEW, 0.3 points or less below the strongest compared method, and 45.66\%/62.27\% on MAFW, ahead of all compared methods; on FERV39k the mean over five splits, 42.30\%/52.66\%, is competitive. The five-fold means are 64.07\%/74.54\% on DFEW, on par with recent large pre-trained models, and 38.90\%/54.37\% on MAFW, below the best of them. A paired $t$-test on DFEW indicates a modest benefit of temporal modeling (+1.61 UAR points, one-sided $p=0.031$). Attention stays on plausible facial regions, and latency is close to a ViT-B/16 baseline (47.4 vs.\ 46.8 ms per clip). Code: \url{https://github.com/ardiytama/GroupMamba-DFER}.
\end{abstract}
 
}
 
\maketitle
 
\IEEEdisplaynontitleabstractindextext
\IEEEpeerreviewmaketitle
 
\section{Introduction}
\label{sec:intro}
\IEEEPARstart{H}{uman} emotion unfolds over time: a spontaneous smile builds from a neutral baseline, peaks, and relaxes, while a flash of disgust may last only a few frames. Static facial expression recognition (SFER) on cropped single frames \cite{ref_poster} discards this temporal structure, which is why dynamic facial expression recognition (DFER) analyzes continuous video instead. Large in-the-wild benchmarks such as FERV39k \cite{ref_ferv39k}, DFEW \cite{ref_dfew}, and MAFW \cite{ref_mafw} have made the task considerably harder by adding unconstrained illumination, camera motion, and occlusion.
 
Two obstacles remain largely unresolved. The first is computational. Spatiotemporal Vision Transformers \cite{ref_vivit,ref_mvit} link distant frames directly, but self-attention scales as $\mathcal{O}(N^2)$ in the number of tokens $N$, forcing a trade-off between clip length, spatial resolution, and GPU memory. Selective state-space models such as Mamba \cite{ref_mamba} match the long-range capacity of attention at $\mathcal{O}(N)$ cost and have been adapted to images \cite{ref_vmamba,ref_visionmamba} and video \cite{ref_videomamba}, but not around the geometric priors specific to facial expression. The second obstacle is label noise. Crowdsourced annotators frequently disagree on ambiguous or transitional expressions \cite{ref_ambiguity,ref_lfa}, and a network trained with plain cross-entropy absorbs this noise, producing the well-documented bias toward majority classes such as ``Happy'' and ``Neutral'' \cite{ref_nrdfernet,ref_ial}.
 
We argue that the two problems should be solved jointly: an efficient backbone without a noise-robust decision layer still overfits corrupted labels, while a robust decision layer on a quadratic backbone still cannot observe enough of the clip. GM-GReFEL couples a linear-complexity spatiotemporal state-space engine with a reliability mechanism grounded in facial geometry. Our contributions are as follows.
\begin{itemize}
\item \textbf{Modulated GroupMamba engine.} Each layer partitions the channels into six groups, scans four of them across the frame and two across time with a Visual Single Selective Scan, and merges them again through a Channel Affinity Modulation gate. Texture and motion thus get separate pathways while the cost of the extractor stays linear in $N$.
\item \textbf{Entropy-guided geometric reliability.} We carry GReFEL \cite{ref_grefel} over from images to video. How uncertain a prediction is, measured by its normalized entropy, decides per sample how much the output relies on the classifier and how much on a cosine-similarity vote against trainable class anchors; under occlusion, blur, or ambiguity the anchors dominate.
\item \textbf{Static-to-Dynamic transfer and Triple Loss.} Spatial weights and anchors learned on still images are inflated into the video model, the temporal parameters start close to zero, and two extra loss terms (anchor orthogonality and center loss) are meant to stop the anchors from collapsing.
\item \textbf{Comprehensive and transparent evaluation.} On its best fold, GM-GReFEL gives the highest UAR and WAR among all compared methods on MAFW and comes within 0.3 points of the strongest method, S4D, on DFEW. We report the five-fold mean next to every best-fold figure: it is comparable to recent pre-trained models on DFEW and lower on MAFW, and the FERV39k result is competitive. The evaluation is supported by a paired $t$-test on DFEW, component and hyperparameter ablations, Grad-CAM and $t$-SNE analyses, and a compute comparison against a ViT-B/16 baseline.
\end{itemize}

\section{Related Work}
\label{sec:related}
 
\subsection{From Image-Based to Video-Based Expression Analysis}
Early DFER systems extracted frame-level features with a 2D CNN and modeled temporal order with a recurrent network, an approach that tracks short-term changes but degrades over long clips because of vanishing gradients. 3D convolutional backbones such as C3D \cite{ref_c3d}, P3D \cite{ref_p3d}, I3D \cite{ref_i3d}, and R(2+1)D \cite{ref_r2plus1d} compute motion within a single tensor operation. Their fixed receptive field, however, captures rapid local motion more readily than slowly forming expressions, and enlarging the kernels raises the parameter count and the risk of overfitting imbalanced in-the-wild collections like DFEW \cite{ref_dfew}, FERV39k \cite{ref_ferv39k}, and MAFW \cite{ref_mafw}.
 
\subsection{Face-Detection-Free DFER Paradigms}
Most DFER pipelines, including ours, assume a face detection and alignment (FDA) stage that crops the face before temporal modeling. PTH-Net \cite{ref_pthnet} questions this assumption by recognizing expressions directly from raw video: a frozen video backbone extracts a temporal feature pyramid at multiple frequencies, and a Temporal Hierarchy Refinement architecture with a Scalable Semantic Distinction layer separates target from non-target frames, preserving body-language cues that cropping discards. Because GM-GReFEL's geometric anchors rely on a stable, aligned facial region, we see the FDA-free paradigm as complementary to ours, and extending the reliability mechanism to unaligned frames is a separate research direction.
 
\subsection{Transformer-Based Spatiotemporal Modeling}
ViViT \cite{ref_vivit} and MViT \cite{ref_mvit} brought global self-attention to video, and Former-DFER \cite{ref_formerdfer} and STT \cite{ref_stt} applied it to facial video. Later work addressed label noise (NR-DFERNet \cite{ref_nrdfernet}), majority-class collapse (IAL \cite{ref_ial}), and multi-instance learning (M3DFEL \cite{ref_m3dfel}); CLIP-based methods (DFER-CLIP \cite{ref_dferclip}, EmoCLIP \cite{ref_emoclip}, A$^3$lign-DFER \cite{ref_a3lign}) inject vision-language priors, and masked-autoencoder methods (MAE-DFER \cite{ref_maedfer}, VideoMAE \cite{ref_videomae}, HiCMAE \cite{ref_hicmae}, SVFAP \cite{ref_svfap}) pursue label-efficient pre-training. Static-to-dynamic methods such as S2D \cite{ref_s2d} and S4D \cite{ref_s4d} transfer image-level facial priors to video, as our initialization does. All of these retain a self-attention core whose $\mathcal{O}(N^2)$ cost limits the resolution and clip length they can process.
 
\subsection{Selective State-Space Backbones in Vision}
Structured state space models propagate an input through a latent state, but early deep-learning variants used fixed transition parameters and could not filter tokens by content. Mamba \cite{ref_mamba} made the step size and projections input-dependent, allowing a linear-time recurrence to discard uninformative tokens while approximating the long-range reasoning of attention. VMamba \cite{ref_vmamba} and Vision Mamba \cite{ref_visionmamba} adapted this recurrence to 2D images through cross-scan and bidirectional scanning, VideoMamba \cite{ref_videomamba} extended it to video, and GroupMamba \cite{ref_groupmamba}, on which we build, partitions channels into groups with distinct scan directions re-coupled by a channel modulation gate. To our knowledge, no prior work combines grouped linear-complexity scanning with an explicit geometric reliability mechanism for facial expression recognition.
 
\subsection{Label Noise and Geometric Reliability}
In-the-wild DFER labels come from crowdsourced annotation, and annotators disagree on subtle or occluded expressions \cite{ref_ambiguity,ref_lfa}. Networks trained on such labels first learn generalizable patterns and later memorize incorrect labels \cite{ref_ambiguity}. Statistical remedies redistribute probability across plausible classes or spread attention across facial regions \cite{ref_lfa}, while center-loss \cite{ref_dacl} and margin-based objectives \cite{ref_elasticface} tighten class clusters; however, all of them correct ambiguity using the same pixel evidence that produced it. GReFEL \cite{ref_grefel} instead introduces an anatomically grounded anchor that is independent of image quality, validating predictions against geometric facial-landmark statistics. We extend this idea from static images to video, replacing a fixed weighting with a per-sample, entropy-based mixing coefficient.
 
\subsection{Summary and Positioning}
Efficient sequence modeling and geometric reliability have so far progressed separately: linear-complexity backbones lack an explicit reliability check, and GReFEL has been formulated only for static images. GM-GReFEL closes this joint gap, using Mamba-style scanning to remove the quadratic cost and letting the network's own predictive entropy decide how much weight the geometric evidence receives.
 
\section{Proposed Method}
\label{sec:method}
 
\subsection{Overview}
\begin{figure*}[!t]
\centering
\includegraphics[width=0.95\textwidth]{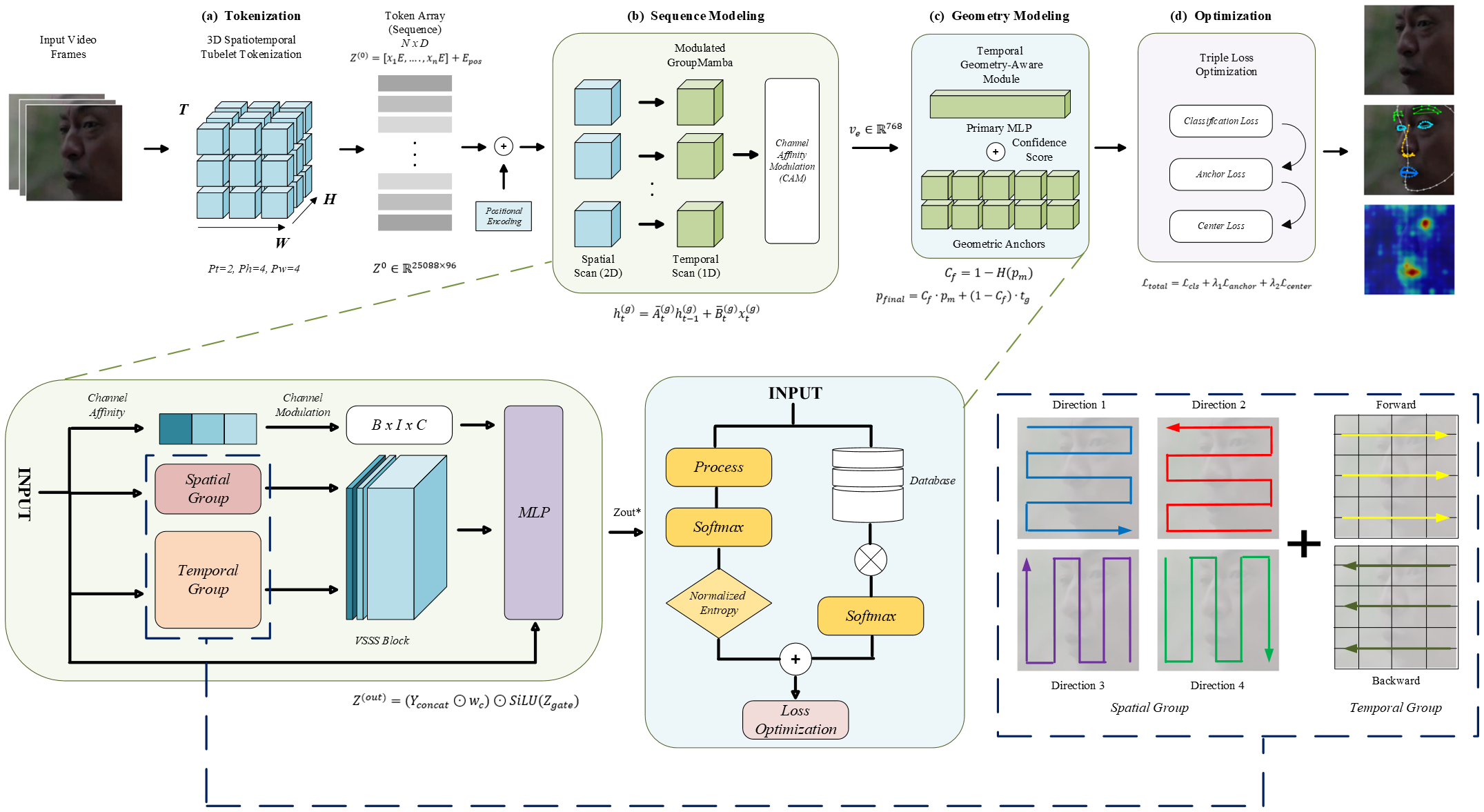}
\caption{Structure of the GM-GReFEL pipeline. Top row: the four-stage forward pass, (a) 3D spatiotemporal tubelet tokenization with positional encoding, (b) sequence modeling via the Modulated GroupMamba engine with Channel Affinity Modulation (CAM), (c) geometry modeling via the Temporal Geometry-Aware module that combines the primary MLP's confidence score with the geometric anchor bank, and (d) Triple Loss optimization, together with example output visualizations (facial landmarks, Grad-CAM attention). Bottom row: internal detail of each stage. Bottom left: the Modulated GroupMamba layer, in which the Spatial Group and Temporal Group each pass through a stack of VSSS blocks before Channel Affinity and Channel Modulation gating produce the $B \times 1 \times C$ layer output $Z^{(out)}$. Bottom center: the Geometry-Aware Reliability mechanism, showing the Statistical Track (Process, Softmax, Normalized Entropy) computed in parallel with the Geometric Track (a similarity lookup against the geometric anchor database via Softmax), combined before Loss Optimization. Bottom right: the six VSSS scanning directions --- four boustrophedon spatial-group sweeps (Direction 1--4) and two temporal-group directions (Forward, Backward).}
\label{fig:architecture}
\end{figure*}

The overall GM-GReFEL pipeline is illustrated in Fig.~\ref{fig:architecture}. A raw video clip is first temporally resampled and spatially normalized (Section~\ref{sec:sampling}), then partitioned into 3D spatiotemporal tubelets and linearly projected into a token sequence (Section~\ref{sec:tokenization}). This sequence is processed by a stack of Modulated GroupMamba layers, in which the channel dimension is split into six groups that are scanned along four spatial directions and two temporal directions using a Visual Single Selective Scan, then re-fused by a Channel Affinity Modulation gate (Section~\ref{sec:groupmamba}). The resulting dense motion-aware feature vector is passed to the Geometry-Aware Reliability module, which computes the network's own predictive entropy and blends the primary classification with a geometrically grounded vote accordingly (Section~\ref{sec:grefel}). All parts are optimized jointly under a Triple Loss objective (Section~\ref{sec:grefel}) after a Static-to-Dynamic weight-transfer initialization phase (Section~\ref{sec:s2d}). Section~\ref{sec:complexity} formalizes the resulting computational complexity.
 
\subsection{Temporal Sampling and Spatial Preprocessing}
\label{sec:sampling}
Two preprocessing choices, frame selection and normalization, measurably affect performance (Section~\ref{sec:ablation}) and are specified here.
 
\subsubsection{Jittered Temporal Sampling}
Clips vary in length (e.g., 45 versus 120 frames), whereas tubelet extraction (Section~\ref{sec:tokenization}) requires $T=16$ frames. At test time we split each clip into $T$ equal segments and take each segment's center frame, which is fully reproducible. Using the same frames at every training epoch, however, would let the temporal groups (Section~\ref{sec:groupmamba}) memorize fixed frame indices, so during training we apply random temporal jittering: with $L_{total}$ available frames, the index selected for step $t \in [0, T-1]$ is
\begin{equation}
I_t = \left\lfloor \frac{L_{total}}{T}\times t + \delta \right\rfloor,
\label{eq:jitter}
\end{equation}
where $\delta$ is drawn per clip and epoch from a discrete uniform distribution over $\left[0, \lfloor L_{total}/T \rfloor - 1\right]$. This perturbs the frame offset within each segment while preserving temporal order, encouraging the temporal transition matrices $\bar{A}_5, \bar{A}_6$ of Eq.~\eqref{eq:recur} to learn motion patterns robust to small sampling shifts; disabling it is the ``w/o Jittered Sampling'' setting in Table~\ref{tab:ablation}.
 
\subsubsection{Spatial Alignment and Normalization}
Before resizing frames to $224\times224$, we detect the face in the first frame and apply the same crop to all $T$ frames, since re-detecting per frame introduces crop jitter that the temporal groups cannot distinguish from real facial motion. Pixel intensities are then standardized per channel,
\begin{equation}
v'_{c,h,w} = \frac{v_{c,h,w} - \mu_c}{\sigma_c}, \qquad c \in \{R,G,B\},
\label{eq:zscore}
\end{equation}
where $\mu_c$ and $\sigma_c$ are computed once over the training split; feeding raw 0--255 intensities into the tubelet projection $E$ of Eq.~\eqref{eq:z0} produced unstable initial gradients.
 
\subsection{Spatiotemporal Tubelet Tokenization}
\label{sec:tokenization}
A video clip is represented as a tensor $\nu \in \mathbb{R}^{T \times C \times H \times W}$, with $T$ sampled frames of $C$ color channels and spatial size $H\times W$. For fair comparison against the datasets used in this study, we resize each frame to $224 \times 224$ and sample $T=16$ frames per clip with $C=3$, giving $\nu \in \mathbb{R}^{16 \times 3 \times 224 \times 224}$.
 
Tokenizing frame by frame would multiply the sequence length and bring back the memory pressure we aim to avoid, so we extract 3D tubelets of size $P_t \times P_h \times P_w$, with $P_t=2$, $P_h=4$, $P_w=4$ in our implementation. This yields
\begin{align}
N_t &= T/P_t = 16/2 = 8, \\
N_h &= H/P_h = 224/4 = 56, \\
N_w &= W/P_w = 224/4 = 56,
\end{align}
and a total token count
\begin{equation}
N = N_t \times N_h \times N_w = 8 \times 56 \times 56 = 25{,}088.
\label{eq:tokencount}
\end{equation}
Setting $P_t=2$ halves the temporal sequence length at the very first layer, which is the primary lever we use to keep the token count, and therefore the downstream memory footprint, tractable for a 16-frame clip. Each flattened tubelet has dimensionality $\mathrm{Dim}(X_i) = P_t \times P_h \times P_w \times C = 2\times4\times4\times3 = 96$, and is projected through a learned matrix $E \in \mathbb{R}^{96 \times D}$ (with $D=96$ at the input stage) and summed with a trainable positional encoding $E_{pos} \in \mathbb{R}^{N \times D}$:
\begin{equation}
Z^{(0)} = [x_1 E, x_2 E, \dots, x_N E] + E_{pos}, \quad Z^{(0)} \in \mathbb{R}^{25{,}088 \times 96}.
\label{eq:z0}
\end{equation}
 
\subsection{Modulated GroupMamba Engine}
\label{sec:groupmamba}
Feeding the $N=25{,}088$-token sequence $Z^{(0)}$ into a self-attention layer would require computing an $N \times N$ correlation matrix, an $\mathcal{O}(N^2)$ operation that is impractical at this scale for standard GPU memory budgets. We instead replace the attention core entirely with a Modulated GroupMamba engine built on a linear, discretized state-space recurrence.
 
\subsubsection{Orthogonal Channel Partitioning}
The channel dimension $D=96$ is evenly divided into $G=6$ groups,
\begin{equation}
Z^{(0)} = [Z_1, Z_2, Z_3, Z_4, Z_5, Z_6], \qquad Z_g \in \mathbb{R}^{25{,}088 \times 16}.
\label{eq:partition}
\end{equation}
The first four groups ($g \in \{1,2,3,4\}$) are dedicated to spatial texture analysis and are scanned across the 2D frame plane along four boustrophedon (serpentine) sweep patterns, illustrated as Direction 1--4 in the bottom-right panel of Fig.~\ref{fig:architecture}, which lets the model capture static structural cues such as wrinkle depth or jawline shape while traversing the full spatial extent of each frame. The remaining two groups ($g \in \{5,6\}$) are permuted so that scanning proceeds exclusively along the temporal axis: Group 5 executes a forward temporal scan and Group 6 a backward temporal scan (also shown in the bottom-right panel of Fig.~\ref{fig:architecture}), so that these pathways track the velocity of a muscle contraction from onset to apex.
 
\subsubsection{Discretized Selective State-Space Recurrence}
\begin{figure}[!t]
\centering
\includegraphics[width=0.85\linewidth]{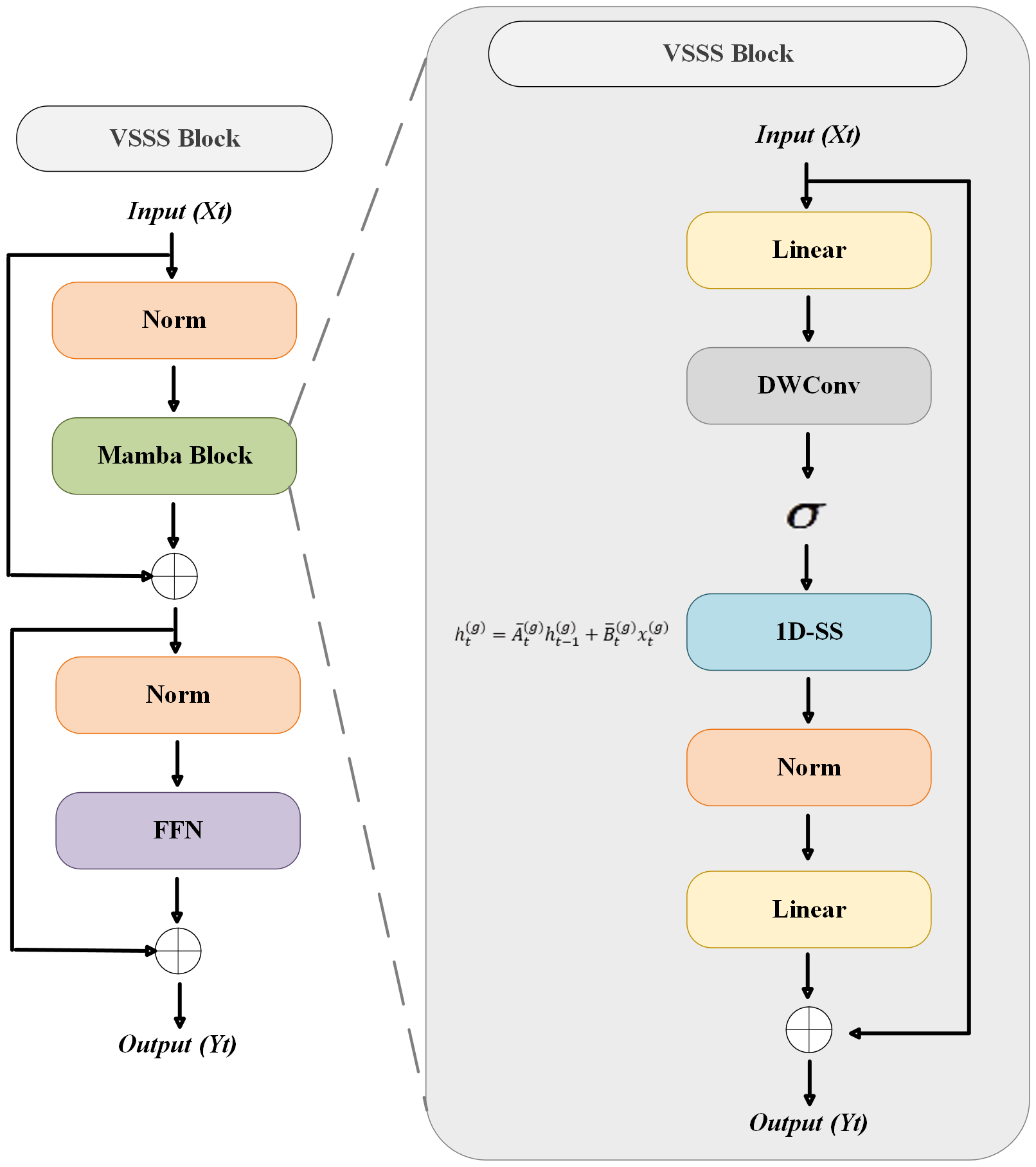}
\caption{VSSS block: a Mamba-style selective scan core and a subsequent feed-forward layer.}
\label{fig:vsss}
\end{figure}

All six sub-tensors $Z_g$ are processed in parallel by VSSS blocks (Fig.~\ref{fig:vsss}). Each block starts from the linear time-invariant system
\begin{align}
h'(t) &= Ah(t) + Bx(t), \label{eq:ssm1}\\
y(t) &= Ch(t), \label{eq:ssm2}
\end{align}
with hidden-state dimension $N_{state}=16$. Because our tokens are discrete, we discretize Eq.~\eqref{eq:ssm1}--\eqref{eq:ssm2} via zero-order hold, and, following the Mamba selectivity mechanism, we make the step size $\Delta_t$ and projection matrices $B_t, C_t$ input-dependent linear functions of the current token $x_t$:
\begin{align}
\Delta_t &= \mathrm{Softplus}(\mathrm{Linear}_\Delta(x_t)), \label{eq:delta}\\
B_t &= \mathrm{Linear}_B(x_t), \qquad C_t = \mathrm{Linear}_C(x_t), \label{eq:bc}\\
\bar{A}_t &= \exp(\Delta_t A), \label{eq:abar}\\
\bar{B}_t &= (\Delta_t A)^{-1}\left(\exp(\Delta_t A) - I\right)\cdot \Delta_t B_t, \label{eq:bbar}\\
h_t^{(g)} &= \bar{A}_t^{(g)} h_{t-1}^{(g)} + \bar{B}_t^{(g)} x_t^{(g)}. \label{eq:recur}
\end{align}
Intuitively, $\Delta_t$ acts as an input-dependent forgetting gate: when the current token carries irrelevant information (e.g., a sudden lighting change or an eye blink), the network drives $\Delta_t$ toward zero, $\bar{A}_t$ approaches the identity matrix, and the hidden state is left largely unchanged, preventing noise from corrupting the accumulated representation of the actual expression.
 
\subsubsection{Channel Affinity Modulation}
Scanning the six groups independently is what keeps the engine linear, but it also leaves the spatial and temporal pathways functionally isolated from one another after each layer. We restore cross-group communication with a Channel Affinity Modulation (CAM) gate, illustrated in the bottom-left panel of Fig.~\ref{fig:architecture}. The six outputs are first concatenated back to the original channel width,
\begin{equation}
Y_{concat} = \mathrm{Concat}(y_{t,1}, \dots, y_{t,6}) \in \mathbb{R}^{25{,}088 \times 96},
\label{eq:concat}
\end{equation}
pooled over all tokens, and mapped by a two-layer MLP to a per-channel weight vector $w_c \in \mathbb{R}^{96}$:
\begin{equation}
w_c = \sigma\left(\mathcal{W}_2\,\delta\left(\mathcal{W}_1\left(\mathrm{GAP}(Y_{concat})\right)\right)\right),
\label{eq:cam}
\end{equation}
with $\sigma$ denoting a sigmoid and $\delta$ a SiLU. The final layer output is obtained by gating $Y_{concat}$ with $w_c$ and a residual SiLU-activated path from the input projection $Z_{gate}$:
\begin{equation}
Z^{(out)} = (Y_{concat} \odot w_c) \odot \mathrm{SiLU}(Z_{gate}).
\label{eq:finalgate}
\end{equation}
After $L$ stacked GroupMamba layers ($L=3$ by default), temporal pooling collapses the sequence into a single dense feature vector $v_e \in \mathbb{R}^{768}$ that summarizes the spatiotemporal signature of the clip.
 
\subsection{Geometry-Aware Reliability Balancing}
\label{sec:grefel}
Relying purely on $v_e$ for classification is risky because in-the-wild datasets are both class-imbalanced and label-noisy, and a network trained by ordinary cross-entropy against such labels tends to overfit the majority classes. We route $v_e$ through a Geometry-Aware Reliable Facial Expression Learning (GReFEL) module that supplies an independent, anatomically grounded check on the primary prediction, structured as two parallel tracks shown in the bottom-center panel of Fig.~\ref{fig:architecture}: a Statistical Track built around the primary MLP classifier, and a Geometric Track that looks up similarity against a database of geometric anchors.
 
\subsubsection{Normalized Entropy as a Confidence Signal}
The primary probability distribution over the $K$ emotion classes is obtained via an MLP and softmax,
\begin{equation}
p_m = \mathrm{Softmax}(\mathrm{MLP}(v_e)) \in \mathbb{R}^K,
\label{eq:pm}
\end{equation}
and its uncertainty is quantified by the normalized Shannon entropy
\begin{equation}
H(p_m) = -\frac{1}{\log(K)}\sum_{k=1}^{K} p_{m,k}\log(p_{m,k}),
\label{eq:entropy}
\end{equation}
which is converted into a confidence score $C_f = 1 - H(p_m) \in [0,1]$.
 
\subsubsection{Geometric Anchor Matching}
In parallel, the module maintains a set of trainable geometric anchors $a_k \in \mathbb{R}^{768}$, one per emotion class, that act as idealized structural prototypes. For instance, the anchor for ``Surprise'' encodes the characteristic spatial signature of widened eyes and an open mouth. The cosine similarity between $v_e$ and every anchor,
\begin{equation}
S_k = \frac{v_e \cdot a_k}{\|v_e\|\|a_k\|},
\label{eq:cossim}
\end{equation}
is converted into a pseudo-target distribution via a temperature-scaled softmax with scale $\tau$:
\begin{equation}
t_g = \mathrm{Softmax}(S/\tau) \in \mathbb{R}^K.
\label{eq:tg}
\end{equation}
 
\subsubsection{Entropy-Weighted Calibration}
The final prediction blends the two signals in proportion to the network's own confidence:
\begin{equation}
p_{final} = C_f \cdot p_m + (1-C_f) \cdot t_g.
\label{eq:pfinal}
\end{equation}
When the primary Mamba engine is confident (low entropy, $C_f \to 1$), the decision is driven almost entirely by $p_m$; when the input is ambiguous or corrupted (high entropy, $C_f \to 0$), the decision shifts toward the geometric anchors $t_g$, so the network falls back on anatomy instead of guessing the majority class.
 
\subsubsection{Triple Loss Formulation}
Training is guided by three complementary objectives. The classification loss is standard cross-entropy against the calibrated prediction,
\begin{equation}
\mathcal{L}_{cls} = -\sum_{k=1}^{K} y_{true,k}\log(p_{final,k}).
\label{eq:lcls}
\end{equation}
To prevent the anchors themselves from collapsing toward one another, which would defeat the purpose of the reliability check, an orthogonality constraint is applied to the anchor matrix $A$ (whose rows are the individual anchors $a_k$):
\begin{equation}
\mathcal{L}_{anchor} = \|AA^T - I\|_F^2.
\label{eq:lanchor}
\end{equation}
Finally, a center-loss term pulls each extracted feature toward its ground-truth anchor, minimizing intra-class variance in the latent space:
\begin{equation}
\mathcal{L}_{center} = \frac{1}{2}\|v_e - a_{y_{true}}\|_2^2.
\label{eq:lcenter}
\end{equation}
The full objective is a weighted sum,
\begin{equation}
\mathcal{L}_{total} = \mathcal{L}_{cls} + \lambda_1 \mathcal{L}_{anchor} + \lambda_2 \mathcal{L}_{center},
\label{eq:ltotal}
\end{equation}
with $\lambda_1, \lambda_2$ tuned so that the geometric terms regularize the primary classification signal without overwhelming it.
 
\subsection{Static-to-Dynamic (S2D) Knowledge Transfer}
\label{sec:s2d}
Video DFER datasets are small relative to static face databases such as AffectNet and RAF-DB, and training the full spatiotemporal architecture from scratch on tens of thousands of clips invites memorization. Masked video pre-training studies also report that uninitialized spatiotemporal models suffer unstable, high-variance early gradients, whereas models inheriting spatial priors converge stably \cite{ref_videomae}. Since facial geometry is the same in a photograph and a video frame, we pre-train a 2D variant on static data and transfer it through a decoupled inflation scheme. The 2D patch-embedding weights $W_{2D} \in \mathbb{R}^{C\times P_h \times P_w \times D}$ are inflated to the 3D tubelet projection $W_{3D}$ by replicating across the temporal stride and normalizing by $P_t$:
\begin{equation}
W_{3D}(t,c,h,w) = \frac{1}{P_t} W_{2D}(c,h,w), \quad \forall t \in [1,P_t].
\label{eq:inflate}
\end{equation}
The four spatial-scanning groups inherit their state-transition parameters from the converged 2D model,
\begin{equation}
\theta_{3D}^{(g)} = \theta_{2D}^{(g)}, \qquad g \in \{1,2,3,4\},
\label{eq:spatialinherit}
\end{equation}
while the two temporal groups, which have no 2D analogue, are initialized near zero, $A_{temporal} \sim \mathcal{N}(0,\sigma^2)$ with $\sigma^2 \approx 10^{-4}$. As a result, the network starts as a purely spatial model and learns to weight motion cues gradually, which is intended to keep early training stable. The anchors converged during 2D pre-training are imported directly, $A_{dynamic}^{(initial)} = A_{static}$, so the reliability module starts from informative anchors in the first dynamic iteration.
 
Optimization has two stages: a warm-up in which the learning rate of inherited spatial parameters and anchors is damped by $\gamma_{freeze}=0.01$ while the temporal matrices train freely, followed by end-to-end fine-tuning.
 
\subsection{Computational Complexity and Inference Behavior}
\label{sec:complexity}
The complexity of a single Modulated GroupMamba block is
\begin{equation}
\mathcal{O}_{\mathrm{GroupMamba}} = 3ND + G \cdot N \cdot D_g \cdot N_{state},
\label{eq:complexity1}
\end{equation}
and since $D_g = D/G$, this reduces to
\begin{equation}
\mathcal{O}_{\mathrm{GroupMamba}} \approx \mathcal{O}(N \cdot D \cdot N_{state}),
\label{eq:complexity2}
\end{equation}
strictly linear in the token count $N$. Doubling the clip length from 16 to 32 frames therefore doubles the compute cost instead of quadrupling it, which allows longer sequences to be processed without the memory wall that limits self-attention baselines.
 
At inference time, ground-truth labels are unavailable, so $\mathcal{L}_{anchor}$ and $\mathcal{L}_{center}$ play no further role and the anchor matrix $A$ is frozen. The forward pass tokenizes the clip, extracts $v_e$ through the frozen GroupMamba stack, computes $p_m$ and its entropy $H(p_m)$, projects $v_e$ against the frozen anchors to obtain $t_g$, and applies Eq.~\eqref{eq:pfinal} to produce $p_{final}$; the arg\,max of $p_{final}$ is the predicted label. Because this calibration relies only on the network's own internal entropy and the frozen anchor bank, it requires no external supervision and adds negligible overhead, which we quantify directly in Section~\ref{sec:efficiency}.
 
\subsection{Evaluation Metrics}
\label{sec:metrics}
Because DFER datasets are heavily class-imbalanced, overall accuracy is misleading: a model that always predicts ``Happy'' can score deceptively well. We therefore report both Weighted Average Recall (WAR), which favors majority classes by construction,
\begin{equation}
\mathrm{WAR} = \frac{\sum_{k=1}^{K}T_k}{\sum_{k=1}^{K}N_k},
\label{eq:war}
\end{equation}
and Unweighted Average Recall (UAR), which averages per-class recall equally regardless of sample count and is the more informative metric for whether the reliability module is actually protecting minority classes,
\begin{equation}
\mathrm{UAR} = \frac{1}{K}\sum_{k=1}^{K}\frac{T_k}{N_k},
\label{eq:uar}
\end{equation}
where class $k$ has $N_k$ test samples, of which $T_k$ are classified correctly.
 
\section{Experiments}
\label{sec:experiments}
 
\subsection{Setup}
\label{sec:setup}
 
\subsubsection{Datasets}
Three in-the-wild benchmarks are used. DFEW \cite{ref_dfew} contains over 16,000 clips taken from films, with severe illumination variation and occlusion. FERV39k \cite{ref_ferv39k} is the largest scene-diverse benchmark currently available and is strongly skewed toward ``Happy'' and ``Neutral''. MAFW \cite{ref_mafw} has 11 compound-emotion classes, including anxiety, helplessness, and disappointment, which overlap in facial geometry with the seven basic emotions. All results are obtained under strict 5-fold cross-validation, with every video functioning as held-out test data in exactly one fold. For DFEW and MAFW we report two numbers. The first is the best of the five cross-validation folds, which shows what the model can reach. The second, the five-fold mean with its standard deviation, shows how stable that result is. As far as we are aware, the baseline figures in our tables are reported as five-fold averages, so the five-fold mean is the like-for-like comparison and the best fold is an upper bound; the comparison tables therefore show both. For FERV39k, the reported figure is the mean over the five splits, for the reason given in Section~\ref{sec:sota}. The checkpoint reported for a fold depends on the dataset, and in every case it is chosen on the evaluation fold itself. For DFEW, the best-fold and five-fold figures come from re-evaluating the saved best checkpoint of each fold with one evaluation script, so the per-class recalls, UAR, and WAR of a fold belong to the same checkpoint. This evaluation is not bit-wise deterministic: repeating it on the Fold~5 checkpoint changed the UAR by 0.35 points and the WAR by 0.18 points (Section~\ref{sec:sota}), which indicates the size of the evaluation noise. For MAFW, each split is taken at its epoch of peak WAR, and for FERV39k the best checkpoint is evaluated at the end of training. The ablation uses each fold at its epoch of peak UAR, with the WAR taken at that epoch. Denoting the metric obtained on fold $i$ as $P_i$, the best-fold result is
\begin{equation}
P_{peak} = \max_{i \in \{1,\dots,5\}} P_i,
\label{eq:kfold}
\end{equation}
and the five-fold statistics reported alongside it are the sample mean $\bar{P} = \frac{1}{5}\sum_{i=1}^{5}P_i$ and sample standard deviation across the same 5 folds.
 
\subsubsection{Implementation Details}
Training and testing used one NVIDIA RTX 3090 GPU (24 GB) in a workstation with an Intel Core i9-12900K, PyTorch 2.0+, and CUDA 11.8. Unless stated otherwise in the ablation studies, the GroupMamba-SSM temporal depth is $L=3$ layers, and the number of epochs trained in each fold is listed in the Supplementary Material. Automatic Mixed Precision (AMP) was used throughout and was necessary at this token count. With FP16 forward computation and an FP32 master copy of the weights to prevent gradient underflow, AMP made it possible to fit a usable batch size into 24~GB of VRAM for the $N=25{,}088$-token sequences of Eq.~\eqref{eq:tokencount}. Training followed the two-stage S2D protocol described in Section~\ref{sec:s2d}: a temporal warm-up phase with spatial parameters and anchors damped by $\gamma_{freeze}=0.01$, followed by full end-to-end fine-tuning, both optimized by AdamW (decoupled weight decay, learning rate $10^{-4}$ decayed by a cosine schedule) with a dropout rate of $0.5$ after the CAM layer.
 
\subsection{Ablation Analysis}
\label{sec:ablation-analysis}
This section examines the contribution of each architectural component, tests the DFEW gain for statistical significance, and characterizes how consistently that performance is achieved across data splits.
 
\subsubsection{Component Ablation}
\label{sec:ablation}
Table~\ref{tab:ablation} gives the DFEW performance of the full model and of five variants, each with one component removed or reduced. All rows come from the same five-fold ablation series and are reported as mean $\pm$ SD over the five folds. They are therefore not comparable with the best-fold headline of Section~\ref{sec:sota}.
 
Removing temporal modeling produces the only clear change: UAR falls from 64.69\% to 63.08\%, a loss of 1.61 points, which is examined further in the paired test below. The other four variants stay within 0.3 points of the full model in mean UAR (from $-0.26$ to $+0.29$). This is small next to the fold-to-fold SD of 1.5 to 2.2 points, so five-fold experiments of this kind cannot separate the contribution of these components from noise. In WAR, no component shows a benefit: the variants score between 0.1 and 1.3 points higher than the full model, again about one SD.
 
Our conclusion is therefore limited. In this setting, temporal modeling is the component with a measurable effect on accuracy. The trainable anchors, the three-layer GroupMamba-SSM stack, the two-stage schedule, and jittered sampling were included for design reasons described in Section~\ref{sec:method}: a geometric reliability check, stable early optimization, and variety in the sampled frames. Their effect on average accuracy is not measurable here, and we do not claim that they improve it. These experiments also do not test whether they help training stability or the best-fold results.
 
\begin{table}[!t]
\renewcommand{\arraystretch}{1.2}
\caption{Component Ablation on DFEW, Mean $\pm$ SD Over Five Folds (Population SD, $n=5$). Each Fold Is Taken at Its Epoch of Highest UAR, With WAR at That Epoch. The Full Model Uses $L=3$}
\label{tab:ablation}
\centering
\begin{tabular}{lcc}
\toprule
Configuration & WAR (\%) & UAR (\%) \\
\midrule
GM-GReFEL (full) & 73.46 $\pm$ 1.21 & 64.69 $\pm$ 1.68 \\
w/o Temporal Modeling & 73.59 $\pm$ 1.28 & 63.08 $\pm$ 1.70 \\
w/o Trainable Anchors (frozen) & 74.69 $\pm$ 1.19 & 64.95 $\pm$ 2.20 \\
GroupMamba-SSM depth $L=1$ & 74.58 $\pm$ 1.49 & 64.94 $\pm$ 1.53 \\
w/ Single-Stage Training & 74.80 $\pm$ 1.37 & 64.40 $\pm$ 2.00 \\
w/o Jittered Sampling & 74.73 $\pm$ 1.52 & 64.68 $\pm$ 1.90 \\
\bottomrule
\end{tabular}
\end{table}
 
\subsubsection{Hyperparameter Sensitivity}
\label{sec:hyperparam-sensitivity}
We swept five training hyperparameters on DFEW Fold~1 under an accelerated 50-epoch schedule, since a full 5-fold sweep at four values each would need 100 training runs. The default configuration ($p=0.5$, $L=3$) appears twice in the sweep and reaches 63.62\% and 61.70\% UAR, so repeated runs of the same configuration differ by about two points, and smaller differences should not be read as effects. Table~\ref{tab:hp_summary} gives the best value for each hyperparameter, and the complete results are in the Supplementary Material. Dropout and the two Triple Loss weights show no clear effect. The temperature $\tau=0.10$ gives the highest UAR (65.20\%), 1.5 to 2.2 points above the other values, which is suggestive but not conclusive. A single GroupMamba layer scores highest in this short schedule (64.19\% UAR), which may reflect faster convergence of the smaller model, and $L=5$ collapses to near-chance accuracy under the S2D initialization. We kept $L=3$, the depth chosen in the initial design of the model, as the default throughout for continuity across experiments. The later sweep and the five-fold ablation of Table~\ref{tab:ablation} show no measurable difference between $L=3$ and $L=1$ (64.69\% against 64.94\% UAR in the ablation, within the fold-to-fold SD), so this choice does not change our conclusions.
 
\begin{table}[!t]
\renewcommand{\arraystretch}{1.15}
\caption{Hyperparameter Sensitivity Summary (DFEW Fold~1, Accelerated Schedule). Differences Below About Two Points Are Within Run-to-Run Variation. Full Sweeps in the Supplementary Material.}
\label{tab:hp_summary}
\centering
\begin{tabular}{lccc}
\toprule
Hyperparameter & Values tested & Best & UAR / WAR (\%) \\
\midrule
Dropout $p$ & 0, 0.25, 0.5, 0.75 & 0.75 & 64.10 / 74.93 \\
Layers $L$ & 1, 3, 5 & 1 & 64.19 / 75.69 \\
$\lambda_2$ (center) & 0, 0.1, 1, 10 & 10 & 64.02 / 75.95 \\
$\lambda_1$ (orth.) & 0, 0.1, 1, 10 & 10 & 64.08 / 75.69 \\
Temperature $\tau$ & 0.01, 0.05, 0.1, 0.5 & 0.10 & \textbf{65.20} / 75.22 \\
\bottomrule
\end{tabular}
\end{table}
 
\subsubsection{Statistical Significance}
\label{sec:significance}
To see whether temporal modeling gives a reliable gain on DFEW, we compared the full model with a spatial-only baseline using a paired $t$-test over the five cross-validation folds. In the baseline, frame-level features are mean-pooled over time and no sequential module is used; two-stage training and selective sampling stay active so that the comparison is fair. The null hypothesis is that temporal modeling brings no UAR benefit. We use a one-sided test because the hypothesis is directional.
 
The full model has the higher UAR in four of the five folds, and the mean paired difference is +1.61 points (SD 1.41; $t(4)=2.56$). The one-sided $p$-value is 0.031, which is below 0.05, so we reject the null hypothesis at that level. The two-sided $p$-value is 0.063, and the 95\% confidence interval for the mean difference is $[-0.14,\,3.36]$ points. The gain is therefore modest, and the evidence for it is moderate, not strong; in Fold~3 the baseline is marginally better ($-0.15$ points). The per-fold values are listed in the Supplementary Material. The full-model mean here (64.69\% UAR) is that of Table~\ref{tab:ablation}; it differs from the DFEW mean in Table~\ref{tab:stability} (64.07\%) because the two use different checkpoint-selection rules (Section~\ref{sec:setup}).
 
The test was run on DFEW only. The five-fold baseline was trained on DFEW alone because five-fold training of the Modulated GroupMamba engine is computationally expensive, so there are no paired results for FERV39k or MAFW and we make no statistical claim for them. With only five folds the test has low power, and we present it as supporting evidence for DFEW and not as a general guarantee.
 
\subsubsection{Cross-Fold Stability}
\label{sec:stability}
The DFEW and MAFW figures in the main comparison tables are the best cross-validation folds, whereas the FERV39k figure is the five-split mean (Section~\ref{sec:sota}). A best fold does not show how reliably a result is reached, so Table~\ref{tab:stability} summarizes all five folds of the same runs by mean and standard deviation; the per-fold values are in the Supplementary Material.
 
On DFEW, the best fold (Fold~5) reaches 66.57\% UAR and 76.46\% WAR, and the five-fold mean is 64.07\% UAR / 74.54\% WAR ($\pm$1.80 / $\pm$1.54). FERV39k shows the tightest spread ($\pm$0.84 / $\pm$0.69); its best split (43.68\% / 53.81\%) lies above the reported five-split mean (42.30\% / 52.66\%). MAFW shows the widest spread ($\pm$5.65 / $\pm$6.35), consistent with an 11-way label space and few samples per class; its Fold~4 (45.66\% / 62.27\% at epoch 38) matches the headline result, while the remaining folds range from 31.51\% to 43.31\% UAR, so the headline value marks the upper end of the fold distribution.
 
\begin{table}[!t]
\renewcommand{\arraystretch}{1.2}
\caption{Cross-Fold Mean and Standard Deviation (Sample SD), DFEW/FERV39k/MAFW}
\label{tab:stability}
\centering
\begin{tabular}{lcccc}
\toprule
Dataset & Mean UAR & Std UAR & Mean WAR & Std WAR \\
\midrule
DFEW & 64.07\% & $\pm$1.80 & 74.54\% & $\pm$1.54 \\
FERV39k & 42.30\% & $\pm$0.84 & 52.66\% & $\pm$0.69 \\
MAFW & 38.90\% & $\pm$5.65 & 54.37\% & $\pm$6.35 \\
\bottomrule
\end{tabular}
\end{table}
 
\subsection{Comparison With Published Methods}
\label{sec:sota}
 
\subsubsection{DFEW}
Table~\ref{tab:dfew} compares GM-GReFEL against a broad set of DFER baselines spanning 3D CNNs, recurrent hybrids, and Transformer architectures. The best fold of our five-fold run (Fold~5) reaches 66.57\% UAR and 76.46\% WAR. The UAR is 9.12 points above the strongest UAR among the older baselines in the table (MIDAS \cite{ref_midas}, 57.45\%, shown rounded to 57.4\%) and 0.23 points below S4D (66.8\% UAR, 76.6\% WAR), the strongest large-scale pre-trained model in Table~\ref{tab:summary}; the WAR is 0.14 points below S4D. Most compared methods are reported as five-fold averages, so the fairer comparison is our five-fold mean of 64.07\% UAR and 74.54\% WAR, given in the last row of Table~\ref{tab:dfew}. On that basis GM-GReFEL is 6.6 points above MIDAS, close to MAE-DFER and VideoMAE (63.4\% and 63.6\% UAR), and 2.7 points below S4D. For both GM-GReFEL rows, the per-class recalls, UAR, and WAR come from one evaluation of the saved best checkpoint of each fold, so the UAR equals the mean of the seven per-class recalls.
 
\begin{table*}[!t]
\renewcommand{\arraystretch}{1.2}
\caption{Recall (\%) of GM-GReFEL and Published Methods on DFEW. \\ H = Happy, S = Sad, N = Neutral, A = Angry, Sp = Surprise, D = Disgust, F = Fear. GM-GReFEL is shown for its best fold (Fold~5) and for the mean over five folds, both from one evaluation of the saved best checkpoints. Bold marks the best value in each column.}
\label{tab:dfew}
\centering
\begin{tabular}{lccccccccc}
\toprule
Method & H & S & N & A & Sp & D & F & UAR & WAR \\
\midrule
C3D \cite{ref_c3d} & 75.1 & 39.4 & 55.1 & 62.4 & 45.0 & 1.3 & 20.5 & 42.7 & 53.5 \\
P3D \cite{ref_p3d} & 74.8 & 43.4 & 54.1 & 60.4 & 50.9 & 0.6 & 23.2 & 43.9 & 54.4 \\
I3D \cite{ref_i3d} & 78.6 & 44.1 & 56.6 & 55.8 & 45.8 & 2.0 & 20.5 & 43.4 & 54.2 \\
R(2+1)D18 \cite{ref_r2plus1d} & 79.6 & 39.0 & 57.6 & 50.3 & 48.2 & 3.4 & 21.0 & 42.7 & 53.2 \\
3D ResNet18 \cite{ref_3dresnet} & 73.1 & 48.2 & 50.5 & 64.7 & 50.1 & 0.0 & 26.3 & 44.7 & 54.9 \\
ResNet18+LSTM \cite{ref_dfew} & 78.0 & 40.6 & 53.7 & 56.8 & 45.0 & 4.1 & 21.6 & 42.8 & 53.0 \\
EC-STFL \cite{ref_dfew} & 79.1 & 49.0 & 57.8 & 60.9 & 46.1 & 2.7 & 21.5 & 45.3 & 56.5 \\
ResNet18+GRU \cite{ref_formerdfer} & 82.8 & 63.8 & 65.0 & 68.5 & 52.0 & 0.8 & 30.1 & 51.6 & 64.0 \\
FormerDFER \cite{ref_formerdfer} & 84.0 & 62.5 & 67.5 & 70.0 & 56.4 & 3.4 & 31.7 & 53.6 & 65.7 \\
STT \cite{ref_stt} & 87.3 & 67.9 & 64.9 & 71.2 & 53.1 & 3.4 & 34.0 & 54.5 & 66.4 \\
DPCNet \cite{ref_dpcnet} & -- & -- & -- & -- & -- & -- & -- & 55.0 & 66.3 \\
NR-DFERNet \cite{ref_nrdfernet} & 88.4 & 64.8 & 70.0 & 75.0 & 61.6 & 0.0 & 19.4 & 54.2 & 68.1 \\
GCA+IAL \cite{ref_ial} & 87.9 & 67.2 & 70.1 & 76.0 & \textbf{62.2} & 0.0 & 26.4 & 55.7 & 69.2 \\
M3DFEL \cite{ref_m3dfel} & 89.5 & 68.3 & 67.8 & 74.2 & 59.6 & 0.0 & 31.6 & 56.1 & 69.2 \\
RDFER \cite{ref_rdfer} & 89.6 & 69.2 & 70.1 & 71.4 & 62.0 & 0.6 & 28.7 & 56.9 & 69.7 \\
ResNet18+TSM MIDAS \cite{ref_midas} & 87.4 & 67.3 & 58.6 & 68.0 & 59.6 & \textbf{28.6} & 44.5 & 57.4 & 69.1 \\
\midrule
\textbf{GM-GReFEL (Ours, best fold)} & \textbf{94.3} & \textbf{85.7} & 70.4 & \textbf{80.7} & 61.4 & 24.1 & \textbf{49.4} & \textbf{66.6} & \textbf{76.5} \\
GM-GReFEL (Ours, 5-fold mean) & 93.1 & 80.5 & \textbf{71.1} & 78.3 & 61.4 & 21.4 & 42.7 & 64.1 & 74.5 \\
\bottomrule
\end{tabular}
\end{table*}
 
Per-fold recall shows that majority classes such as ``Happy'' are essentially bulletproof across folds (91.2--94.9\%), while minority classes like ``Disgust'' show substantially more inter-fold variance, which we attribute to the scarcity of samples in those splits (full per-fold breakdown in the Supplementary Material).
 
\begin{figure}[!t]
\centering
\includegraphics[width=\linewidth]{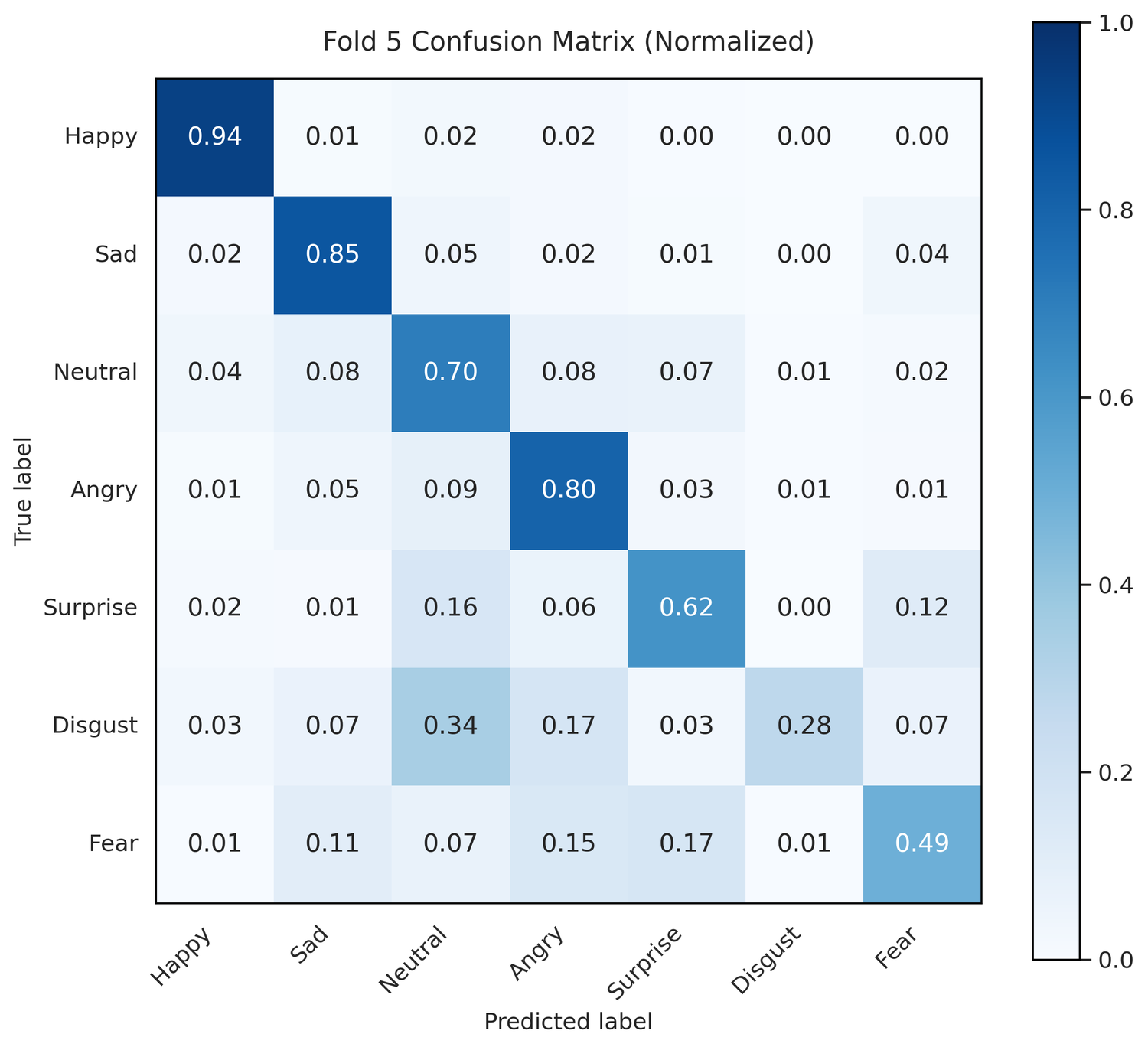}
\caption{Normalized confusion matrix for DFEW, Fold~5 (repeated evaluation of the saved checkpoint; see text).}
\label{fig:dfew_confmat}
\end{figure}

In the best fold, the recall is 94.3\% on Happy, 85.7\% on Sad, and 80.7\% on Angry, while the minority classes Fear (49.4\%) and Disgust (24.1\%) remain the weakest (Table~\ref{tab:dfew}). Fig.~\ref{fig:dfew_confmat} shows the normalized confusion matrix of the Fold~5 checkpoint from a repeated evaluation. Because the evaluation is not bit-wise deterministic, the repetition differs slightly from Table~\ref{tab:dfew}: the UAR changes by 0.35 points, the WAR by 0.18 points, and the largest change in a single recall is 3.4 points, which is one of the 29 Disgust test clips (8 instead of 7 correct). We therefore use the figure only to show the pattern of confusions. Misclassifications follow facial anatomy: ``Disgust'' is confused with ``Neutral'' (34\%) and ``Angry'' (17\%), and ``Fear'' bleeds into ``Surprise'' (17\%), since both expressions widen the eyes in a similar way.
 
\subsubsection{FERV39k}
FERV39k presents an even more severe class-imbalance challenge than DFEW, with minority emotions such as ``Disgust'' and ``Fear'' constituting a small fraction of the dataset. Table~\ref{tab:ferv} compares GM-GReFEL against the strongest published baselines. Among the methods in Table~\ref{tab:ferv}, the strongest UAR was 35.94\% (M3DFEL \cite{ref_m3dfel}); GM-GReFEL raises this to 42.30\%, a +6.36-point gain, with WAR reaching 52.66\%. These figures are means over the five splits of our cross-validation run, not the best split (43.68\% / 53.81\%, Split~3). We report the mean because the splits agree closely: the coefficient of variation of UAR is 1.99\% for FERV39k, against 2.82\% for DFEW and 14.52\% for MAFW. With this little variation, the mean is what a further held-out split would be expected to reproduce. Several large-scale pre-trained ViT-B/16 models in Table~\ref{tab:summary} remain ahead on this dataset, as discussed in the cross-dataset summary below. Per-split results are in the Supplementary Material and Section~\ref{sec:stability}.
 
\begin{table}[!t]
\renewcommand{\arraystretch}{1.2}
\caption{UAR and WAR (\%) of GM-GReFEL and Published Methods on FERV39k}
\label{tab:ferv}
\centering
\begin{tabular}{lcc}
\toprule
Method & WAR & UAR \\
\midrule
C3D \cite{ref_c3d} & 31.69 & 22.68 \\
P3D \cite{ref_p3d} & 33.39 & 23.20 \\
I3D \cite{ref_i3d} & 38.78 & 30.17 \\
R(2+1)D18 \cite{ref_r2plus1d} & 41.28 & 31.55 \\
3D ResNet18 \cite{ref_3dresnet} & 37.57 & 26.67 \\
R18+LSTM \cite{ref_ferv39k} & 42.95 & 30.92 \\
2R18+LSTM \cite{ref_ferv39k} & 43.20 & 31.28 \\
Former-DFER \cite{ref_formerdfer} & 45.72 & 36.88 \\
NR-DFERNet \cite{ref_nrdfernet} & 45.97 & 33.99 \\
GCA+IAL \cite{ref_ial} & 48.54 & 35.82 \\
M3DFEL \cite{ref_m3dfel} & 47.67 & 35.94 \\
\midrule
\textbf{GM-GReFEL (Ours, 5-split mean)} & \textbf{52.66} & \textbf{42.30} \\
\bottomrule
\end{tabular}
\end{table}
 
FERV39k's training dynamics follow the same pattern as DFEW, with smooth validation loss decay and no catastrophic forgetting. Its confusion matrix shows the same anatomy-driven error structure: ``Disgust'' and ``Fear'' scatter toward classes with overlapping muscular signatures (e.g., ``Fear'' bleeding into ``Sadness''), which suggests that the anchors degrade gracefully under data scarcity (full curves and confusion matrix in the Supplementary Material).
 
\subsubsection{MAFW}
MAFW tests a different axis of difficulty: 11-way compound-emotion classification, where categories such as ``Disappointment'' and ``Anxiety'' are geometrically close to more basic states. On its best fold, GM-GReFEL reaches 45.66\% UAR and 62.27\% WAR (Table~\ref{tab:mafw}), which is 3.83 points higher in WAR and 1.94 points higher in UAR than the strongest baseline, S4D \cite{ref_s4d} (58.44\% WAR, 43.72\% UAR). MAFW also has the widest spread across folds. The five-fold mean is 38.90\% UAR and 54.37\% WAR ($\pm$5.65 and $\pm$6.35), shown in the last row of Table~\ref{tab:mafw}. On this basis GM-GReFEL has a lower UAR than T-MEP, DFER-CLIP, and the self-supervised models (39.4--43.7\%) and a WAR similar to SVFAP and MAE-DFER (54.28\% and 54.31\%).
 
\begin{table}[!t]
\renewcommand{\arraystretch}{1.2}
\caption{UAR and WAR (\%) of GM-GReFEL and Published Methods on MAFW}
\label{tab:mafw}
\centering
\begin{tabular}{lcc}
\toprule
Method & WAR & UAR \\
\midrule
Former-DFER \cite{ref_formerdfer} & 43.27 & 31.16 \\
T-MEP \cite{ref_tmep} & 52.85 & 39.37 \\
DFER-CLIP \cite{ref_dferclip} & 52.55 & 39.89 \\
EmoCLIP \cite{ref_emoclip} & 41.46 & 34.24 \\
A$^3$lign-DFER \cite{ref_a3lign} & 53.24 & 42.07 \\
SVFAP \cite{ref_svfap} & 54.28 & 41.19 \\
HiCMAE \cite{ref_hicmae} & 56.17 & 42.65 \\
MAE-DFER \cite{ref_maedfer} & 54.31 & 41.62 \\
S2D \cite{ref_s2d} & 57.37 & 41.86 \\
VideoMAE \cite{ref_videomae} & 53.51 & 40.87 \\
S4D \cite{ref_s4d} & 58.44 & 43.72 \\
\midrule
\textbf{GM-GReFEL (Ours, best fold)} & \textbf{62.27} & \textbf{45.66} \\
GM-GReFEL (Ours, 5-fold mean) & 54.37 & 38.90 \\
\bottomrule
\end{tabular}
\end{table}
 
MAFW's training dynamics show strong generalization, with validation loss settling below training loss. Its confusion matrix separates structurally distinct emotions cleanly (Disgust, Fear, Disappointment, and Anger all above 78\% true-positive rate) while ambiguous compound classes such as ``Surprise'' and ``Anxiety'' scatter toward anatomically adjacent basic emotions, so even the errors follow facial physiology (full curves, confusion matrix, and per-class breakdown in the Supplementary Material).
 
\subsubsection{Cross-Dataset Summary}
Table~\ref{tab:summary} lists the compared methods on all three benchmarks, with GM-GReFEL shown for its best fold and as a five-fold mean. On the best fold, GM-GReFEL has the highest UAR and WAR among all listed methods on MAFW and comes within 0.3 points of S4D on DFEW (66.6\% against 66.8\% UAR, 76.5\% against 76.6\% WAR). The five-fold mean is the like-for-like basis for most baselines, and on it the picture is more balanced. On DFEW, GM-GReFEL (64.1\% UAR, 74.5\% WAR) is level with the recent pre-trained ViT-B/16 models (61.8--64.0\% UAR, 74.2--76.0\% WAR) and below S4D (66.8\%, 76.6\%). On MAFW, its mean UAR (38.9\%) is below that of T-MEP, DFER-CLIP, and the self-supervised models (39.4--43.7\%), while its mean WAR (54.4\%) is close to SVFAP and MAE-DFER. On FERV39k, three pre-trained models (MAE-DFER, VideoMAE, and S4D) report higher UAR (43.1--43.4\% versus our 42.3\%), and S4D also has the higher WAR (53.6\% versus 52.7\%). Averaged over the three datasets and both metrics, our headline figures give 57.7\% against 57.1\% for S4D, whereas the five-fold means give 54.5\%. We therefore regard GM-GReFEL as competitive with large pre-trained models but not uniformly ahead of them.
 
\begin{table*}[!t]
\renewcommand{\arraystretch}{1.2}
\caption{UAR and WAR (\%) on DFEW, FERV39k, and MAFW, Grouped by Method, Sampling, and Backbone. DS = Dynamic Sampling; TI = Time Interpolation; GroupMamba-T = the tiny-configuration GroupMamba backbone used throughout this paper (Section~\ref{sec:groupmamba}). For FERV39k, both GM-GReFEL rows give the five-split mean. Bold marks the best value in each column.}
\label{tab:summary}
\centering
\begin{tabular}{lcccccc}
\toprule
\multirow{2}{*}{Method (Sampling)/Backbone} & \multicolumn{2}{c}{DFEW} & \multicolumn{2}{c}{FERV39k} & \multicolumn{2}{c}{MAFW} \\
\cmidrule(lr){2-3} \cmidrule(lr){4-5} \cmidrule(lr){6-7}
& UAR & WAR & UAR & WAR & UAR & WAR \\
\midrule
EC-STFL (TI)/C3D+P3D \cite{ref_dfew} & 45.3 & 56.5 & -- & -- & -- & -- \\
Former-DFER (DS)/Trans. \cite{ref_formerdfer} & 53.6 & 65.7 & 37.2 & 46.8 & 31.1 & 43.2 \\
CEFLNet (CLIP)/ResNet-18 \cite{ref_ceflnet} & 51.1 & 65.3 & -- & -- & -- & -- \\
NR-DFERNet (DS)/CNN-Trans. \cite{ref_nrdfernet} & 54.2 & 68.1 & 33.9 & 45.9 & -- & -- \\
STT (DS)/ResNet-18 \cite{ref_stt} & 54.5 & 66.6 & 37.7 & 48.1 & -- & -- \\
EST (DS)/ResNet-18 \cite{ref_est} & 53.9 & 65.8 & -- & -- & -- & -- \\
Freq-HD (FreqHD)/VGG13-LSTM \cite{ref_freqhd} & 46.8 & 55.6 & 33.0 & 45.2 & -- & -- \\
LOGO-Former (DS)/ResNet-18 \cite{ref_logoformer} & 54.2 & 66.9 & 38.2 & 48.1 & -- & -- \\
IAL (DS)/ResNet-18 \cite{ref_ial} & 55.7 & 69.2 & 35.8 & 48.5 & -- & -- \\
AEN (DS)/ResNet-18 \cite{ref_aen} & 56.6 & 69.3 & 38.1 & 47.8 & -- & -- \\
T-MEP (DS)/Transformer \cite{ref_tmep} & 57.1 & 68.8 & -- & -- & 39.3 & 52.8 \\
M3DFEL (DS)/ResNet-18-3D \cite{ref_m3dfel} & 56.1 & 69.2 & 35.9 & 47.6 & -- & -- \\
CLIPER (DS)/CLIP-ViT-B/16 \cite{ref_cliper} & 57.5 & 70.8 & 41.2 & 51.3 & -- & -- \\
DFER-CLIP (DS)/CLIP-ViT-B/32 \cite{ref_dferclip} & 59.6 & 71.2 & 41.2 & 51.6 & 39.8 & 52.5 \\
EmoCLIP (DS)/CLIP-ViT-B/32 \cite{ref_emoclip} & 58.0 & 62.1 & 31.4 & 36.1 & 34.2 & 41.4 \\
A$^3$lign-DFER (DS)/CLIP-ViT-L/14 \cite{ref_a3lign} & 64.0 & 74.2 & 41.8 & 51.7 & 42.0 & 53.2 \\
SVFAP (DS)/ViT-B/16 \cite{ref_svfap} & 62.8 & 74.2 & 42.1 & 52.2 & 41.1 & 54.2 \\
HiCMAE (DS)/ViT-B/16 \cite{ref_hicmae} & 63.7 & 75.0 & -- & -- & 42.6 & 56.1 \\
MAE-DFER (DS)/ViT-B/16 \cite{ref_maedfer} & 63.4 & 74.4 & 43.1 & 52.0 & 41.6 & 54.3 \\
S2D (DS)/ViT-B/16 \cite{ref_s2d} & 61.8 & 76.0 & 41.2 & 52.5 & 41.8 & 57.3 \\
VideoMAE (DS)/ViT-B/16 \cite{ref_videomae} & 63.6 & 74.6 & 43.3 & 52.3 & 40.8 & 53.5 \\
S4D (DS)/ViT-B/16 \cite{ref_s4d} & \textbf{66.8} & \textbf{76.6} & \textbf{43.4} & \textbf{53.6} & 43.7 & 58.4 \\
\midrule
\textbf{GM-GReFEL, best fold (DS)/GroupMamba-T (Ours)} & 66.6 & 76.5 & 42.3 & 52.7 & \textbf{45.7} & \textbf{62.3} \\
GM-GReFEL, 5-fold mean (DS)/GroupMamba-T (Ours) & 64.1 & 74.5 & 42.3 & 52.7 & 38.9 & 54.4 \\
\bottomrule
\end{tabular}
\end{table*}
 
\subsection{Visualization}
\label{sec:qualitative}
 
\subsubsection{Training Convergence}
\label{sec:convergence}
\begin{figure}[!t]
\centering
\includegraphics[width=\linewidth]{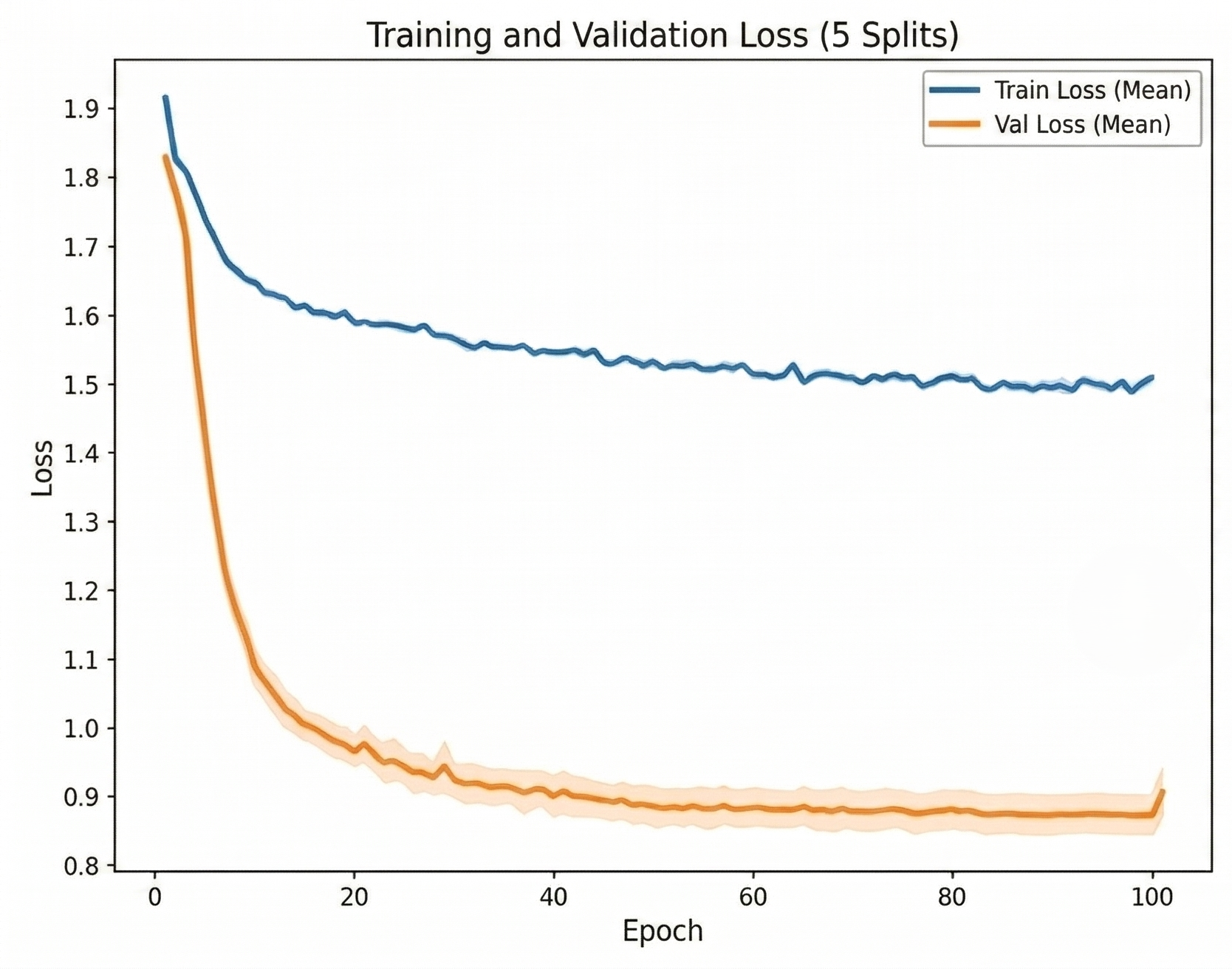}
\caption{Mean training and validation loss trajectory on DFEW, averaged across all five cross-validation folds with shaded variance bands.}
\label{fig:dfew_loss}
\end{figure}

Because DFEW is representative of the general unconstrained-video setting, we use it to illustrate convergence behavior; the same qualitative pattern held across all three datasets. Fig.~\ref{fig:dfew_loss} plots the mean training and validation loss across all five folds, with shaded variance bands. Training loss stabilizes at a visibly higher plateau ($\approx 1.5$) than validation loss ($\approx 0.9$); this is the expected consequence of the aggressive 50\% dropout applied after the CAM layer during training, which is disabled at validation time. The narrow variance bands across folds indicate that this convergence pattern is not an artifact of any single data split.
 
\begin{figure}[!t]
\centering
\includegraphics[width=\linewidth]{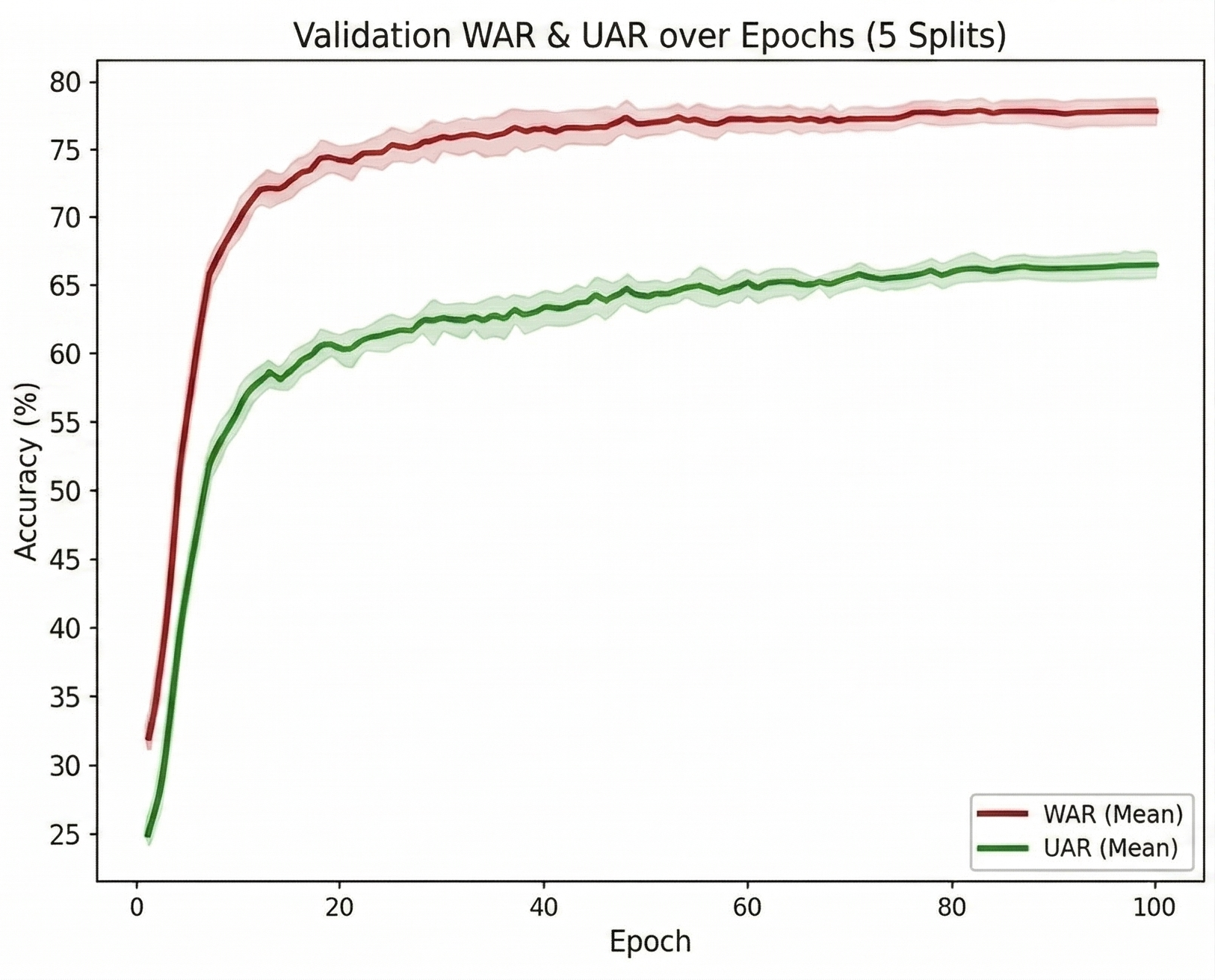}
\caption{Progression of UAR and WAR on DFEW across training, averaged across five folds.}
\label{fig:dfew_uarwar}
\end{figure}

UAR and WAR both track a consistent trajectory across training, as shown in Fig.~\ref{fig:dfew_uarwar}. Both metrics rise sharply during the first 20 epochs and then settle into a stable plateau. The narrow variance bands over the final 50 epochs show that training converges consistently across folds.
 
\subsubsection{Interpretability and Attention Analysis}
\begin{figure}[!t]
\centering
\includegraphics[width=\linewidth]{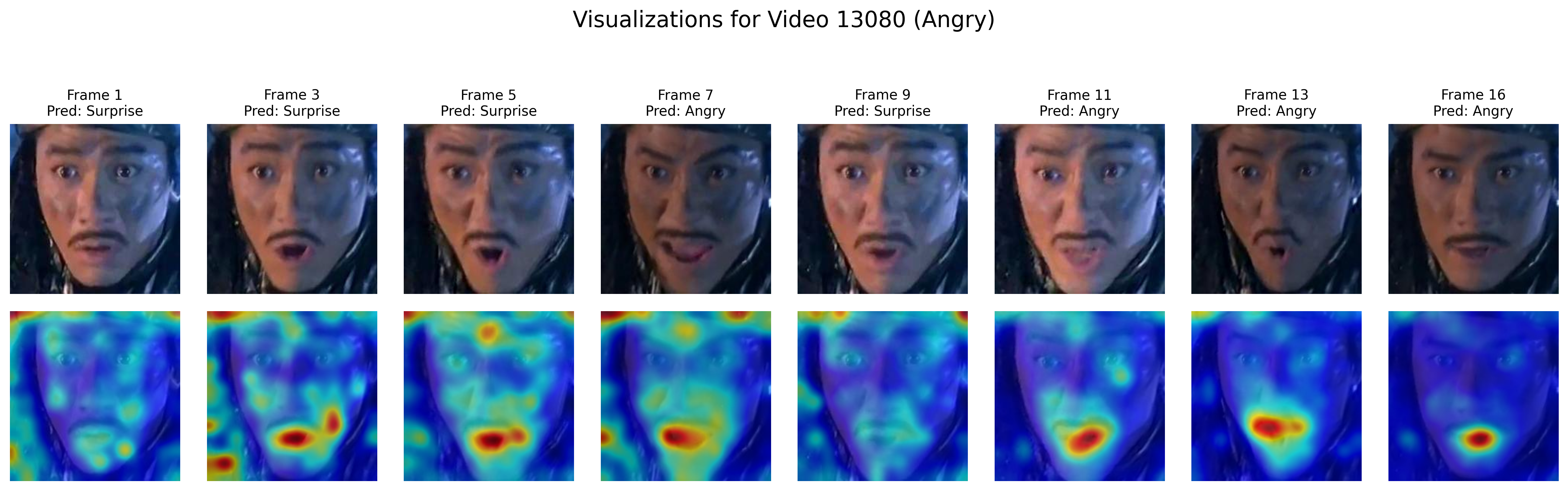}
\caption{Grad-CAM attention across a DFEW clip labeled ``Angry,'' showing attention shifting from the ocular region toward the mouth and jawline as the expression reaches its apex.}
\label{fig:gradcam_angry}
\end{figure}

To assess whether GM-GReFEL relies on facial physiology or on incidental background cues, we applied Grad-CAM to the final Modulated GroupMamba layer and projected the resulting gradients back onto the input frames. Fig.~\ref{fig:gradcam_angry} traces attention across a representative DFEW clip labeled ``Angry.'' In the earliest frames, attention concentrates on the ocular region as the subject shows a surprise-like widening of the eyes; as the temporal groups detect the characteristic downward brow motion and mouth tightening that follow, attention shifts decisively toward the mouth and jawline by the emotional apex. This suggests that the network follows the temporal evolution of the expression instead of classifying frames in isolation. These are illustrative case studies, not statistical evidence; four further case studies are in the Supplementary Material.
 
\begin{figure}[!t]
\centering
\includegraphics[width=\linewidth]{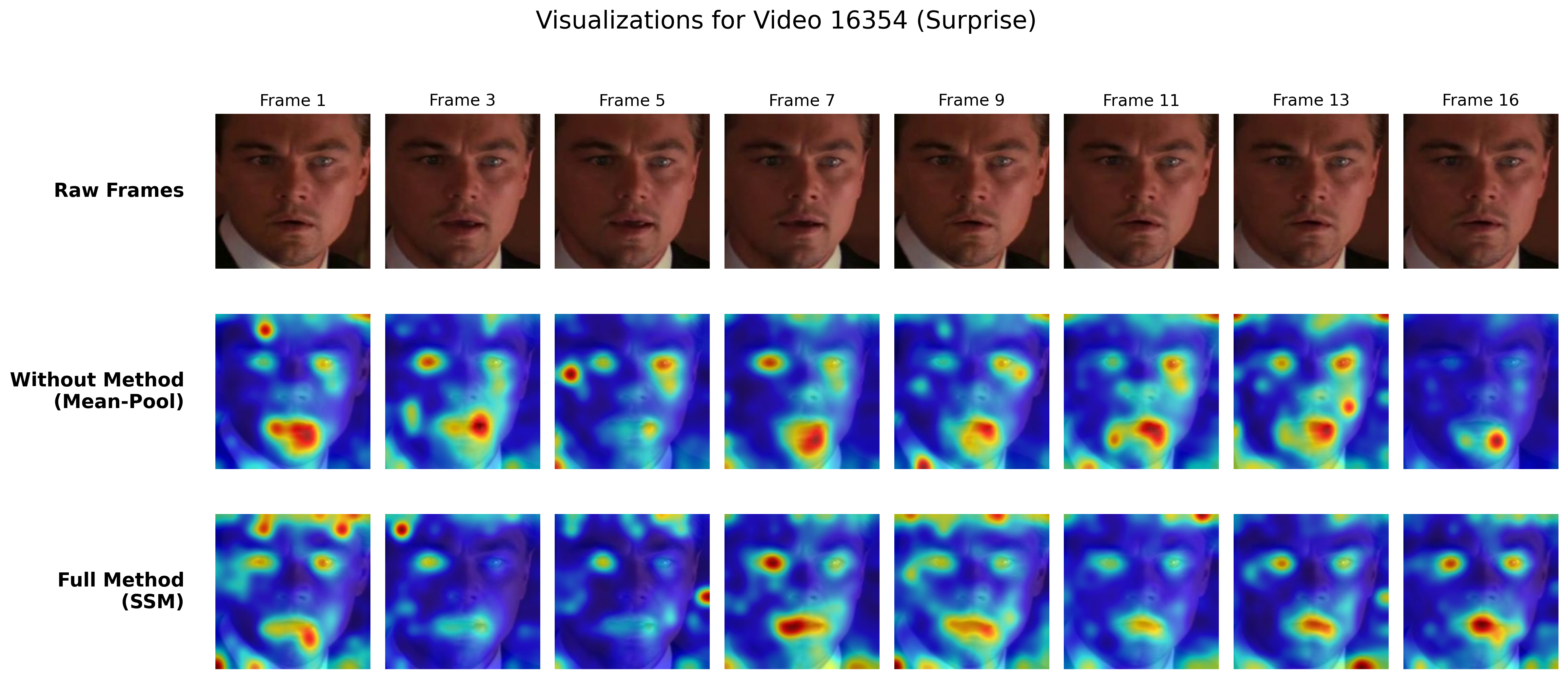}
\caption{Comparison of attention maps for a ``Surprise'' sample: full GroupMamba-SSM model (localized on mouth and eyes) vs.\ a mean-pooling baseline (diffuse, drifting into background).}
\label{fig:gradcam_compare}
\end{figure}

A direct side-by-side comparison makes the effect concrete: Fig.~\ref{fig:gradcam_compare} places attention maps from the full GroupMamba-SSM model against a mean-pooling baseline on the same ``Surprise'' sample. The baseline's attention is diffuse, bleeding into hair, clothing, and background; the full model's attention stays tightly localized on the mouth and eyes and remains stable across the 16-frame sequence. This may reflect the constraint that the geometric anchors place on attention, although the ablation in Section~\ref{sec:ablation} does not isolate their effect on accuracy.
 
\subsubsection{t-SNE Feature Space Analysis}
\begin{figure}[!t]
\centering
\includegraphics[width=\linewidth]{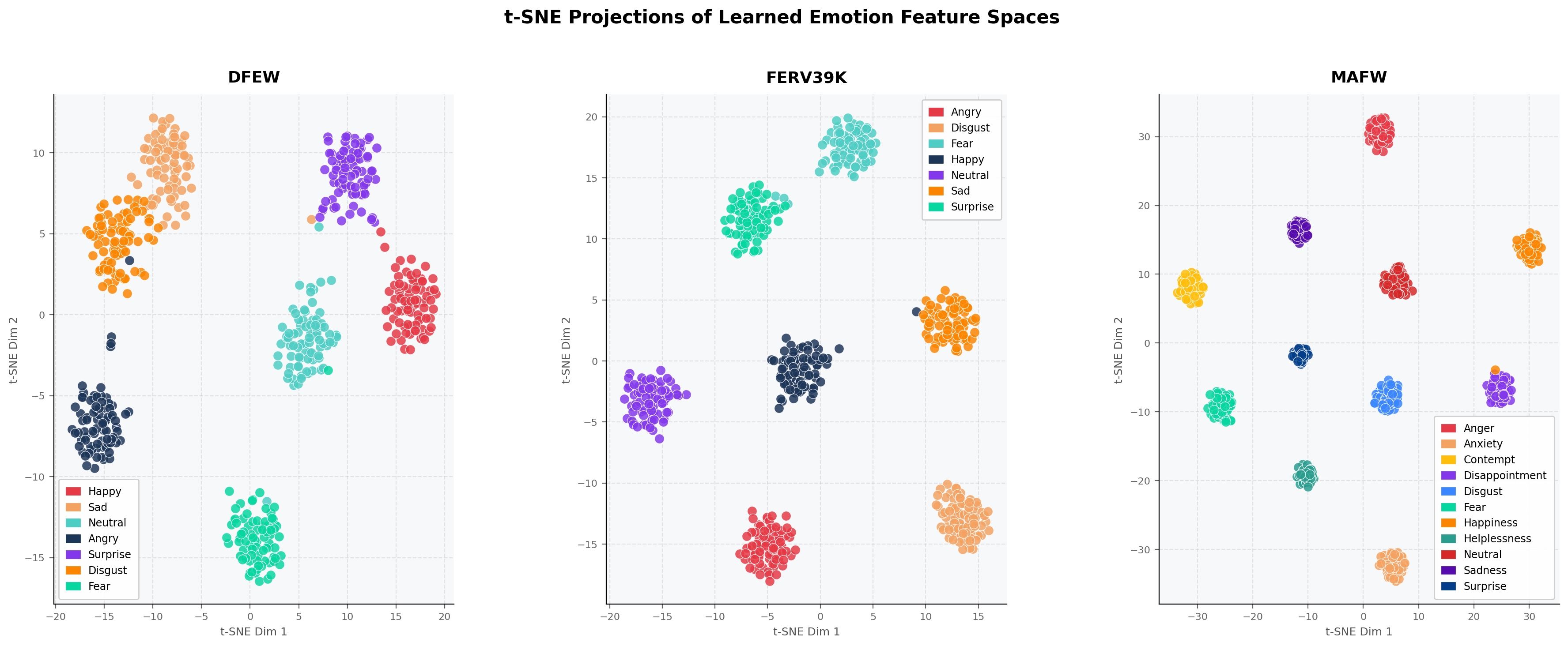}
\caption{$t$-SNE projections of the learned 768-dimensional latent feature space for DFEW, FERV39k, and MAFW.}
\label{fig:tsne}
\end{figure}

Projecting the learned 768-dimensional feature space into two dimensions via $t$-SNE (Fig.~\ref{fig:tsne}) offers a complementary, global view of separability. In each case, the architecture forms compact, well-separated clusters with minimal inter-class overlap. Even the 11 compound classes of MAFW remain separable, which agrees with the Triple Loss objective reducing intra-class variance and enlarging inter-class margins in the shared latent space.
 
\subsubsection{Confusion and Misclassification Patterns}
A radar comparison of per-class recall shared between FERV39k and MAFW, a breakdown of the most frequent misclassification pairs, cross-fold WAR/UAR stability across the full 5-fold cycle, and a comparison of per-class recall against F1-score are given in the Supplementary Material. On FERV39k, seven of the ten most frequent errors involve the ``Neutral'' class, so the main remaining difficulty there appears to be locating the onset and offset of an expression. On MAFW the frequent errors are spread over pairs of related compound emotions, such as Neutral to Fear, Happiness to Contempt, and Disappointment to Disgust, and only three of the ten involve ``Neutral''.
 
\subsection{Computational Efficiency}
\label{sec:efficiency}
Table~\ref{tab:efficiency} compares GM-GReFEL with a ViT-B/16 baseline profiled under identical settings. The reliability module raises the parameter count to 155.99M versus 147.69M (+5.6\%), but replacing quadratic self-attention with the linear GroupMamba engine lowers computation to 539.62 versus 548.75 GFLOPs. Measured throughput is essentially unchanged (21.1 versus 21.4 clips/s; 47.4 versus 46.8 ms per clip), so the reliability machinery does not compromise practical inference speed.
 
\begin{table}[!t]
\renewcommand{\arraystretch}{1.2}
\caption{Computational Complexity and Inference Speed Comparison}
\label{tab:efficiency}
\centering
\begin{tabular}{lccc}
\toprule
Metric & GM-GReFEL & ViT-B/16 & $\Delta$ \\
\midrule
Total Params & 155.99 M & 147.69 M & +8.30 M (+5.6\%) \\
GFLOPs & 539.62 & 548.75 & $-9.13$ \\
FPS (clips/s) & 21.1 & 21.4 & $-0.3$ ($-1.4\%$) \\
Latency (ms) & 47.4 & 46.8 & +0.6 \\
\bottomrule
\end{tabular}
\end{table}
 
Table~\ref{tab:literature-efficiency} places GM-GReFEL alongside other DFER methods. Only the GM-GReFEL and ViT-B/16 rows were profiled locally (224$\times$224, $T=16$, \texttt{thop}, RTX 3090); the remaining figures are taken from each method's publication and may use different input settings, so they serve as an approximate reference.
 
\begin{table}[!t]
\renewcommand{\arraystretch}{1.2}
\caption{Compute Comparison With Other DFER Methods. $^\dagger$: Measured Locally (224$\times$224, $T{=}16$, \texttt{thop}, RTX 3090). Other Rows Are Taken From Each Method's Publication and May Use Different Settings; -- Means Not Reported.}
\label{tab:literature-efficiency}
\centering
\begin{tabular}{lccc}
\toprule
Method & Backbone & Params (M) & GFLOPs \\
\midrule
M3DFEL \cite{ref_m3dfel} & ResNet-18-3D & 37.5 & 58.0 \\
SVFAP-B \cite{ref_svfap} & ViT-B/16 & 78.0 & 44.0 \\
MAE-DFER \cite{ref_maedfer} & ViT-B/16 & 85.0 & 50.0 \\
VideoMAE \cite{ref_videomae} & ViT-B/16 & 86.0 & 82.0 \\
S4D \cite{ref_s4d} & ViT-B/16 (MoAE) & 90.0 & -- \\
A$^3$lign-DFER \cite{ref_a3lign} & CLIP-ViT-L/14 & -- & -- \\
ViT-B/16 baseline$^\dagger$ & ViT-B/16 & 147.69 & 548.75 \\
\textbf{GM-GReFEL (Ours)}$^\dagger$ & GroupMamba-T & \textbf{155.99} & \textbf{539.62} \\
\bottomrule
\end{tabular}
\end{table}
 
GM-GReFEL has more parameters than every listed method, including the ViT-B/16-based SVFAP-B, MAE-DFER, VideoMAE, and S4D (78--90M) and the much lighter M3DFEL. This capacity resides in the GroupMamba-SSM engine and the reliability module, whose contributions are examined in Section~\ref{sec:ablation-analysis}; the case for GM-GReFEL rests on its accuracy and on the linear-complexity argument of Section~\ref{sec:complexity}, not on parameter count. GFLOPs for S4D and A$^3$lign-DFER are not reported in their publications and are left blank; we did not estimate them.
 
\section{Limitations}
\label{sec:limitations}
Three scope constraints bound this study. First, GM-GReFEL is unimodal: it uses visual input only, whereas emotion is also conveyed by voice and gesture, and aligning audio streams with the tubelet schedule is outside the present scope. Second, it performs categorical recognition and does not model dimensional (valence--arousal) affect, because the geometric anchors are tied to discrete class prototypes and cannot directly track blended or graded affect. Third, although the backbone is linear in sequence length, the full pipeline still requires a laboratory-grade GPU to reach the latency in Table~\ref{tab:efficiency}; quantization and pruning for edge deployment remain future work.
 
\section{Conclusion}
\label{sec:conclusion}
This paper presented GM-GReFEL, a DFER architecture that jointly addresses the quadratic memory cost of self-attention and the label noise of in-the-wild datasets. A Modulated GroupMamba engine scans orthogonal spatial and temporal channel groups with a linear state-space recurrence and re-couples them through Channel Affinity Modulation, and an entropy-guided geometric reliability module provides an anatomical fallback whenever the statistical prediction is uncertain. On its best fold, GM-GReFEL gives the highest UAR and WAR among the compared methods on MAFW and comes within 0.3 points of the strongest method on DFEW. Its five-fold means are comparable to recent pre-trained models on DFEW and lower on MAFW, and its FERV39k result is competitive. On DFEW, removing temporal modeling costs a small but consistent amount of accuracy according to a paired test. The attention maps stay on the face, the emotion clusters in feature space are well separated, and inference is about as fast as with a ViT-B/16 baseline. Future work includes fusing audio and visual state-space streams, modeling dimensional affect with a continuous anchor manifold, and compressing the model for real-time deployment.
 
\section*{Acknowledgment}
We are grateful to the CIHCI Lab for discussions and feedback on this work, and to NTUST for the GPUs used in the experiments.

\begin{IEEEbiographynophoto}{Yudhistira Arditya Pratama}
received the B.S. degree in Mechatronics Engineering from Politeknik Elektronika Negeri Surabaya (PENS), Surabaya, Indonesia, and the M.S. degree in Electrical Engineering from National Taiwan University of Science and Technology (NTUST), Taipei, Taiwan. He is currently pursuing the Ph.D. degree in Electrical Engineering at NTUST. His research interests include computer vision, dynamic facial expression recognition, spatiotemporal state-space modeling, and underwater image processing.
\end{IEEEbiographynophoto}

\begin{IEEEbiographynophoto}{Yi-Zeng Hsieh}
(Senior Member, IEEE) received the B.S., M.S., and Ph.D. degrees in computer science and information engineering from National Central University, Taoyuan City, Taiwan, in 2004, 2006, and 2012, respectively. He is currently a Professor with the Department of Electrical Engineering, National Taiwan University of Science and Technology, Taipei, Taiwan. He has authored or coauthored more than 100 journal and refereed conference papers. His research interests include deep learning, machine learning, neural networks, pattern recognition, image processing, computer vision, and swarm intelligence. He is a Senior Member of both the IEEE Consumer Technology Society and the IEEE Signal Processing Society, and was recognized as an Outstanding Young Scholar by the IEEE Consumer Technology Society in 2023.
\end{IEEEbiographynophoto}

\end{document}